# Robust Optimization of Unconstrained Binary Quadratic Problems

Mark Lewis*
Craig School of Business, Missouri Western State University, Saint Joseph, MO, 64507, USA
*Corresponding author

Gary Kochenberger
School of Business, University of Colorado, 1250 14th St., Denver, CO, 80202, USA

John Metcalfe
Department of Computer Science and Engineering, University of Kansas, Lawrence, KS, 66045, USA

**Abstract**. In this paper we focus on the unconstrained binary quadratic optimization model, maximize $x^t Q x$, x binary, and consider the problem of identifying optimal solutions that are robust with respect to perturbations in the Q matrix.. We are motivated to find robust, or stable, solutions because of the uncertainty inherent in the big data origins of Q and limitations in computer numerical precision, particularly in a new class of quantum annealing computers. Experimental design techniques are used to generate a diverse subset of possible scenarios, from which robust solutions are identified. An illustrative example with practical application to business decision making is examined. The approach presented also generates a surface response equation which is used to estimate upper bounds in constant time for Q instantiations within the scenario extremes. In addition, a theoretical framework for the robustness of individual $x_i$ variables is considered by examining the range of Q values over which the $x_i$ are predetermined.

**Keywords**: Robust Optimization; Unconstrained Binary Quadratic Problems; Upper Bounds; Business Decision Making; Scenario Generation; Experimental Design; Surface Response Equation; Sensitivity Analysis

**Biographical Notes**: Mark Lewis has a BS in Electrical Engineering from the University of Kansas and worked as an engineer in avionic development at Lockheed Martin for twelve years before earning a PhD in Operations Research from Southern Methodist University in Dallas, Texas, USA. He is currently a professor in the Masters of Information Management program at Missouri Western State University. He has published in International Journal of Operational Research, INFORMS Journal on Computing, Computers and Operations Research, Operations Research Letters, Networks, European Journal of Operational Research and others.

John Metcalfe is a student in Computer Science and Electrical Engineering at the University of Kansas. He is a software engineer who has developed 3D gaming architectures and software designs. He has expertise in multiple computer languages.

Gary Kochenberger's academic specialty concerns Building, Testing, and Implementing Algorithms for solving resource allocation problems that arise in both the private and public sectors. In recent years, his focus has been on problems of a combinatorial nature. He has published four books and numerous research papers in such journals as the Operational Research Management Science, Mathematical Programming, International Journal of Operational Research, Journal of Optimization Theory and Applications, Operation Research, Computers and Operations research, Naval Research Logistics

Quarterly, Decision Sciences, Journal of the Operational Research Society, European Journal of OR, Interfaces, Operations Research Letters, Omega, and the Journal of the Production and Operations Management Society.

## 1 Introduction

Combinatorial optimization models have a lengthy history of both important application and of computational challenge. Applications are reported in every conceivable industry as exemplified by recent work on airport gate assignments (Gunaneskaran, et. al., 2016), school bus routing (Saeid Samadi-Dana et. al., 2016), and resource constrained project scheduling (Joshi et. al., 2016). Many other examples exist as reported in this and other journals.

A combinatorial model of particular importance here is the Unconstrained Binary Quadratic Optimization Problem (UBQP) due to its breadth of application as detailed in the recent survey paper (see Kochenberger et al., 2014). Formally, UBQP is defined as:

$$\text{maximize } x^t Q x \quad \text{s.t. } x \in \{0,1\}$$

where Q is a symmetric *n x n* matrix of real, or integer values. An equivalent form of UBQP is

$$\text{maximize } x^t Q x = \sum_i c_{ii} x_i + \sum_{\substack{i,j \\ i \neq j}} c_{ij} x_i x_j \text{ where } x_i \in \{0,1\} \tag{1}$$

which highlights that the elements of the Q matrix consist of the diagonal linear coefficients $c_{ii}$ and the off-diagonal quadratic coefficients $c_{ij}$. Thus all the problem data exists in the Q matrix.

Solution stability in light of data error is major concern in all applications. In this paper we discuss the robust optimization of UBQP where the objective of our work is to investigate the effects of Q matrix perturbations on solution stability. This work is further motivated by recent advances in quantum computing (Choi, 2008) that rely on problems being modeled as UBQP and which are providing exciting opportunities for UBQP research. In addition, quantum annealing computers cannot set the values of the Q matrix with the precision of typical computers, hence the interest in the robustness of solutions to either imprecise or variable Q elements. As a side note, a benefit of studying robustness of UBQP is that this model has no constraints and hence are always feasible, so that modelers can focus solely on variability in Q versus having to also consider variability in constraint parameters and issues of feasibility.

Although UBQP is NP-hard, exact solvers such as Cplex have been shown to be effective for small to medium sized edge sparse problems (currently up to about 1000 variables and 1% density). A common approach is to leverage powerful linear programming techniques and data structures by converting the quadratic objective elements to linear constraints. However, as the size and density of Q increases, this approach creates very large linear programming problems. As problem size and Q density increase, metaheuristic techniques such as path relinking (Wang et al., 2012) are providing fast high quality solutions.

The UBQP model is known to represent many different problem types including Ising problems and a large number of NP problems such as graph and number partitioning, covering and set packing, satisfiability, matching, spanning tree as well as others (Lucas, 2014). The application potential of UBQP can be greatly expanded to include many constrained models by employing reformulation techniques as outlined in the survey article (Kochenberger G. , et al., 2014). Utilizing these methods many models not originally cast in the form of UBQP can be easily reformulated into this unified framework. This approach has been applied to the maximum clique problem (Bomze et al., 1999), max-cut (Boros & Hammer, 1991), scheduling (Alidaee et al., 1994) and many other problem types. The broad range of applications of the UBQP model implies that analyzing robustness as we explore in this paper has direct implications for such analysis for many combinatorial optimization problems.

Recent progress with meta-heuristics has pushed the size of solvable UBQP to more than 15000 decision variables and millions of interactions (Lewis, Kochenberger, & Alidaee, A new modeling and solution approach for the set-partitioning problem, 2008). Small to medium sized UBQP can now be solved to optimality very quickly via various excellent heuristics as well as commercial software (Kochenberger G. et al., 2013; Benoist et al., 2011), so that robust optimization based on solving and analyzing a large number of scenarios is a reasonable technique to investigate.

Robust optimization, as defined in (Mulvey et al., 1995; Libura, 2010), generates a series of solutions to various scenarios where all possible realizations of weights Q create a set of scenarios. A solution that is *close to optimal* for all scenarios is termed "solution robust" and those that are *almost feasible* for all scenarios are "model robust". Unconstrained BQP are inherently model robustness, and we take advantage of this in the approach we report on in this paper by focusing on solution robustness.

As the size and density of Q increases, the number of scenarios to consider when analyzing robustness becomes unmanageably large. As a result, the analysis must be carried out on a carefully constructed subset of scenarios. In this paper the structured subset of scenarios is created using an experimental design approach that generates on the order of 2*n* scenarios. Optimizing this set of scenarios creates a corresponding set of optimal solutions and objective values that are used as a measure of robustness to changes in Q.

A comprehensive survey of robust optimization, including Taguchi's robust design methodology and experimental design, is presented in (Beyer & Sendhoff, 2007) and its use in software testing is detailed in (Dunietz et al., 1997). The application of Taguchi concepts to mixed integer problems such as network design can be found in (Lewis, 2008). Using mixed integer solvers such as Cplex and Gurobi for exactly solving UBQP along with comparisons to heuristics is discussed in (Billionet & Elloumi, 2007; Lewis & Kochenberger, 2013).

Our work here focuses on finding optimal solutions to UBQP which are robust to perturbations in the elements of the Q matrix. Upper bound generation, variable fixing via preprocessing and sensitivity analysis are discussed in Section 2. A mathematical model of scenario generation using orthogonal arrays is presented in Section 3. In Section 4 computational results using benchmark UBQP test problems illustrate that the benchmark optimal solution is not robust to minor perturbations in Q and in Section 4.1 we show that small, realistic business problems present analysis challenges. Section 5 presents a discussion of how the robust optimization results can be used to estimate a tight upper bound on the maximum value of $x^t Q x$ when faced with random perturbations in Q.

## 2. Variable Fixing and Sensitivity Analysis

It is often the case when working with data from natural phenomenon that the inherent random component is treated by setting $c_{ij} \in Q$ as an average of a given probability distribution. Similarly large emprical data sets, for example data from measuring flows between nodes in a network to be used in a maximum cut problem, allow calculating $c_{ij} \in Q$ either as minimums, maximums, or more commonly averages. Regardless of how the $c_{ij} \in Q$ are calculated, being able to quickly calculate a best case objective value is useful when a large number of Q scenarios are to be evaluated since solvers may spend considerable time returning an initial solution value and as the size of Q increases, the number of scenarios to evaluate and hence the time needed also increases. A simple upper bound for an instance of UBQP can be quickly calculated by summing all positive elements of $Q$. maximize $x^t Q x \leq \sum_{i,j} c_{ij}$ where $c_{ij} > 0$.

Simple rules, presented below, may be employed to improve this bound by fixing certain variables, yielding a modified Q matrix. The concept of persistency, variable assignments that are valid in some or all optimal solutions, is discussed in (Boros etal., 2006) which preprocesses Q via solving maximum flow problems. Preprocessing of the Q matrix can reduce the upper bound when it is determined that $x_i = 0$ so that any positive interactions involving $x_i$ can be subtracted from the upper bound as explained in Rule 2 below. In section 5 we compare this approach to that based on the surface response equation and show the latter bound is much tighter. The preprocessing rules described below help to motivate the discussion of robustness in terms of a sensitivity analysis. The question being, if a variable $x_k$ is predetermined, then over what range of values $c_{kj}$ and $c_{jk}$ does $x_k$ stay predetermined?

Rules for reducing the size of Q by setting some $x_i$ to 0 or 1 prior to optimization, without affecting optimality are based on the concept of diagonal dominance.

**Rule 1**. For $c_{ii} > 0$ set diagonally *dominant* $x_i = 1$

where diagonally dominant is defined as (for simplicity of notation all Q matrices are symmetric, hence the doubling of the off-diagonal coefficients):

$$\text{For a given } i \text{ where } c_{ii} > 0, \text{ if } c_{ii} + 2 \sum_{\substack{i,j \\ i \neq j}} c_{ij} > 0 \text{ where } c_{ij} < 0 \text{ then } x_i = 1 \qquad (2)$$

in the optimal solution, and the row and column information for $x_i$ can be removed from the Q matrix to create a reduced $Q$. In other words, if a positive diagonal element is larger in magnitude than the sum of the associated negative off-diagonal elements, then it will be set to 1 in the optimal solution, regardless of the other variable values. An obvious corollary is that if all elements of a given i are non-negative, then $x_i$ will be one in the optimal solution.

$$\text{Rule 1a. For a given } i \text{ where } c_{ii} \geq 0, \text{ if } \forall j \; c_{ij} > 0 \text{ then } x_i = 1 \text{ in the optimal solution} \qquad (2a)$$

**Rule 2**. for $c_{ii} < 0$ set diagonally *recessive* $x_i = 0$

where diagonally *recessive* is defined as:

For a given $i$ such that $c_{ii} < 0$, if $c_{ii} - 2\sum_{\substack{i,j \\ i \neq j}} c_{ij} < 0$ where $c_{ij} > 0$ then $x_i = 0$ (3)

in the optimal solution, and the row and column information for $x_i$ can be removed from the Q matrix to create a reduced $Q$. Similar to (1a), if all i are negative, then $x_i$ will be zero in the optimal solution.

Rule 2a. For a given $i$ where $c_{ii} \leq 0$, if $\forall j\ c_{ij} < 0$ then $x_i = 0$ in the optimal solution (3a)

## 2.1 Sensitivity analysis

The rules presented above can be used to provide a measure of robustness via a sensitivity analysis of Q by determining the range of values over which elements of the solution $x$ are predetermined. Let $\Delta p_i$ equal the change in $c_{ii}$, as defined by (4) below, that if positive indicates the amount $c_{ii}$ can be reduced Q and $x_i = 1$. If a given $i$ satisfies Rule 1 then

$$\Delta p_i = c_{ii} + 2\sum_{\substack{ij \\ i \neq j}} c_{ij} \text{ where } c_{ij} < 0 \text{ and } c_{ii} > 0 \quad (4)$$

If $\Delta p_i < 0$ then it is the amount $c_{ii}$ must be increased before $x_i$ can be predetermined equal to one.

For example, in the Q matrix in Figure 1 below, variable $x_3 = 1$ in the optimal solution because $100 - 92 \geq 0$ and Eq (4) states that if $c_{3\ 3}$ drops by 8 then $x_3$ can no longer be assumed to be equal to one in the optimal solution.

**Figure 1**       Example Q matrix with predetermined $x_3 = 1$

|    | x0  | x1  | x2   | x3  | x4   |
|----|-----|-----|------|-----|------|
| x0 | 50  | -75 | 50   | -15 | 0    |
| x1 | -75 | 100 | 0    | -5  | 0    |
| x2 | 50  | 0   | 100  | -25 | -120 |
| x3 | -15 | -5  | -25  | 100 | -1   |
| x4 | 0   | 0   | -120 | -1  | 100  |

Equations (2) and (4) also provide the amount decrease $\Delta c_{mn}$ in each individual $c_{mn}$ that keeps $x_i = 1$ for those $x_i$ satisfying Rule 1. For example, in the problem above $x_3$ will be set to one and $c_{2,3} = -25$. $\Delta c_{2,3} = 100 + 2\ (-25 - 15 - 5\ - 1) = 8$ so that if $c_{2,3}$ decreases by 4 to –29, then $x_3$ can no longer be guaranteed to be equal to one in the optimal solution.

Similarly the range of $c_{ii}$ that will allow a variable satisfying Rule 2 to stay set to zero can also be developed. Conversely (4) provides the amount $c_{ii}$ must increase to allow predetermination. Thus, decision makers analyzing a UBQP model can flag for special attention those $x_i$ that are close to being set to either 0 or 1. The number of variables that may be fixed by this approach is dependent on the problem type and structure of the Q matrix.

## 3    Scenario Generation and Analysis Methodology

Realistically, the coefficients of a model are expected to change or are not known with precision. Scenario generatation solves a set of possible scenarios and then analyzes the results. However, it is very likely that there will be an extremely large set of scenarios, making a total enumeration approach an intractable problem. For example, a relatively sparse UBQP with 100 coefficients, each with a low and high level, would have $2^{100}$ possible scenarios to consider. Thus, testing a subset of possible scenarios is the reasonable option with the question being how best to generate this subset?

The scenario generation technique used in this paper is based on 2-level fractional factorial design concepts employing orthogonal arrays to create diverse unbiased samples. Using this approach, the number of scenarios generated is on the order of the number of elements of Q, so that a relatively sparse UBQP with 100 variables and 1000 Q elements would generate approximately 1000 scenarios. As UBQP grow in size and density they become more difficult to solve exactly, hence efficient metaheuristic techniques would be used to solve a large number of large problems. Quantum annealing computers have demonstrated they can solve large UBQP in microseconds and are another promising avenue of research.

Two extreme scenarios, Q $^{upper}$ and Q $^{lower}$ known as the *scenario generators* define the upper and lower bounds for each element of *Q*. Each test run, or row, in the experimental design represents a scenario instantiated by a Q matrix. Each scenario is optimized and the optimal solution and objective value stored, with the resultant set of optimization data analyzed for robustness. The process is described in Figure 2.

In Figure 2, lines (1) and (2) are inputs defining the best and the worst case values for each $Q_{ij}$ such that all scenarios generated consist of combinations of elements from these two matrices. The number of scenarios *k* to be generated in line (3) is based on the cardinality of $D = \{ (i,j) : Q_{ij}^{upper} <> Q_{ij}^{lower} , \forall (i,j) \in Q\}$ so that *D* is the set of differing elements. Let $d = |D|$. The experimental design *E* in line (4) is a table consists of *k* rows (scenarios) and *d* columns defining the high/low settings of the individual Q elements of a scenario. In other words, *E* is a *k* by *d* matrix where each row $E^i$ is used to create a scenario $Q^i \in \tilde{Q}$ the set of all generated scenarios. Define $\tilde{Q} = \{ Q^k : Q_{ij}^k$ is determined by $E^k$ for all rows $k \in E, (i,j) \in D\}$, that is, for each $(i,j)$ element of *D*, $E_{ij}^k$ determines whether $Q_{ij}^k$ is set to $Q_{ij}^{upper}$ or $Q_{ij}^{lower}$ according to the methods described in (Pan et al., 2008).

Lines (5-8) indicate each scenario generated via *E* creates an instance of UBQP to be solved. Note that in line (5) the order is not important, each test run is independent, thus the loop is easily parallelizable. Define the set of optimal solutions from the experimental design as $\tilde{x} = \{x: xQ^ix = \max xQ^ix,$ for all $Q^i \in \tilde{Q}\}$.

**Figure 2** Robust Optimization Pseudocode

1. $Q^{upper} \leftarrow$ Define upper limits for all $(i,j) \in Q$
2. $Q^{lower} \leftarrow$ Define lower limits for all $(i,j) \in Q$

3. $k \leftarrow$ Calculate_number_of_scenarios ( $d$ )
4. $E \leftarrow$ Create_experimental_design ($Q^{upper}$, $Q^{lower}$, $D$, $k$)

5. for $i = 1$ to $k$
6.     $Q^i \leftarrow$ Create_scenario ($E^i$, $Q^{upper}$, $Q^{lower}$)
7.     $x \leftarrow$ Solve_UBQP ($Q^i$)
8.     $\tilde{x} \leftarrow \tilde{x} \cup x$

## 4    Computational Testing

The experimentally designed scenario generation approach was tested on a subset of the Beasley OR-Library UBQP problem set (Beasley, 1990) to determine their robustness to minor perturbations in $Q$. The ten benchmark UBQP with $|n| = 50$ variables were solved optimally using Cplex v12.5 on a Dell personal computer with 16 GB RAM and 3.4 GHz 4-core hyper-threaded i7-2600 processor running Windows 7. These small problems were readily solved in an average of one second per problem.

To test the effects of changes in $Q$, the $Q_{ij}^{upper}$ and $Q_{ij}^{lower}$ values were set at ±5% of the average values in the benchmark problems. Scenarios were generated, optimized and the results recorded. Summary results presented in Table 1 indicate that the optimal solution for 4 out of 10 problems was unaffected by the ±5% changes in Q because there was only one optimal solution for all scenarios. Of the remaining six problems, four with multiple optima still had the majority of the scenarios covered by the one solution identified as the average – the one associated with the benchmark. Two problems, numbers 1 and 10, had multiple optima with the average solution covering less than a third of the scenarios. Thus, a relatively small change in $Q_{ij}$ led to the creation of multiple optima for six of the ten problems.

It is expected that as the magnitude of the difference in the elements of the scenario generators increases, the diversity of the individual scenarios $Q \in \tilde{Q}$ will increase, and the number and diversity of optimal solutions $x \in \tilde{x}$ will increase. For example, problem 2 in Table 1 found the same solution for all 256 scenarios when $Q^{upper}$ and $Q^{lower}$ were set at ±5% of $Q^{average}$, but when based on ±10% there were six optimal solutions with the $Q^{average}$ solution now covering only 50% of the scenarios. Conversely, the more tightly bound the scenarios, the more likely a dominant robust solution will exist and if there is only one scenario, unless you specifically search for and find alternate optima, there will be only one solution.

**Table 1** Results of Robust Optimization on Benchmark Problems

| Problem # | Number Scenarios tested | Number unique optimal solutions | % of scenarios covered by benchmark solution |
|---|---|---|---|
| 1 | 256 | 4 | 25% |
| 2 | 256 | 1 | 100% |
| 3 | 512 | 1 | 100% |
| 4 | 256 | 1 | 100% |
| 5 | 512 | 6 | 69% |
| 6 | 256 | 1 | 100% |
| 7 | 256 | 6 | 50% |
| 8 | 512 | 2 | 75% |
| 9 | 512 | 6 | 84% |
| 10 | 256 | 13 | 32% |

*4.1      Pedagogical Illustration using Business Simulations*

A small, but reasonably complex example is presented to illustrate that even small problems can become complex. Business simulations with realistic modeling of competitive and environmental forces are a popular pedagogical approach to teaching decision making (International Collegiate Business Strategy Competition, 2014; Leger, 2006). In the ERPSim Meuslix Manfucturing simulation (Leger, 2006), teams of students manage companies producing and selling various cereal products in a competitive environment. Decisions made include: changing product design, stopping production of a product, marketing, pricing, loan management, production capacity, and production efficiencies.

Suppose a company feels that one of its products is changing from the growth to the maturity stage in its product life cycle and wants to analyze possible options for this transition. After a review of historical data on products with similar product life cycles, nine options to be considered were couched as yes/no type decisions as described in Table 2.

The team is interested in estimating the long term (strategic) and short term (tactical) effects of the decisions on net income. They derive two sets $Q^{upper}$ (strategic) and $Q^{lower}$ (tactical) with 9 main effects (diagonal elements) and 14 interactions shown in Figure 3, thus the two matrices in Figure 3 are the scenario generators used to create the set of scenarios $\tilde{Q}$. Note that interaction penalties equal to -100 prevent opposing decisions from both being selected in the same solution, such as decision numbers 0 and 1. A DOE generator that was used in (Lewis & Kochenberger, 2016) to generate the probability of a variable being set to 0 or 1, also allows for the creation of very large fractional factorial experimental designs. Based on having 23 coefficients subject to variance, 64 orthogonal scenarios were generated..

**Table 2**          Nine decisions to analyze for their effects on net income

| Decision # | Description of decision |
|---|---|
| 0 | Enhance Product 1 |
| 1 | Phase out Product 1 and do not replace it, pursue other products and markets |
| 2 | Immediately discontinue Product 1, sell Product 1 resources, switch to other Products and markets |
| 3 | Replace Product 1 with Product X |
| 4 | Market Product 1 |
| 5 | Market Product X |
| 6 | Pursue International market for Product 1 |
| 7 | Outsource production of Product 1 |
| 8 | Pursue domestic cost efficiencies in production of all Products, e.g. ERP, Automation |

**Figure 3**          Net Income Scenario Generators for Table 2

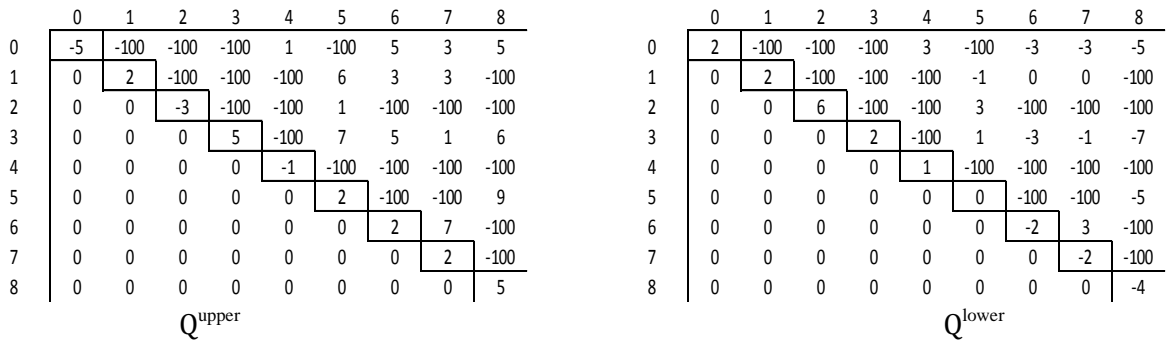

$Q^{upper}$                                                              $Q^{lower}$

All 64 scenarios are solved exactly via Cplex and the set of solutions $\tilde{x}$ along with their average optimal objective function value and frequency is shown in Table 3. For example, the solution $x_4$ occurred 22 times out of the 64 scenarios and the average of the optimals was 22. $x_4$ occurred the most often as the optimal solution and qualifies as the most robust solution. Note that this solution corresponds to: Replace Product 1 with Product X, Market Product X, and Pursue cost efficiencies in production. Alternatively, using $Q^{avg}$, the optimal solution is $x_8$ = Enhance Product 1, Pursue International Market, and Outsource Product 1 which is obviously a completely different implementation. This example shows that the robust optimization approach lends legitimacy to the decision making process in ways that using a simple average cannot.

The results are also helpful in grouping similar decisions. For example, solution $x_3$ is similar to solution $x_4$, differing only in option 8 "Pursue cost efficiencies in production" and these two similar decisions together cover half of the 64 scenarios. Such information is extremely useful for decision making.

**Table 3**    Set of solutions $\tilde{x}$ found during experimental design

| Solution | Average Optimal Value | Frequency of Occurence |
|---|---|---|
| $x_0 = 000001001$ | 15 | 4 |
| $x_1 = 000100001$ | 13 | 3 |
| $x_2 = 000100110$ | 13 | 2 |
| $x_3 = 000101000$ | 12 | 10 |
| $x_4 = 000101001$ | 22 | 22 |
| $x_5 = 001001000$ | 9 | 1 |
| $x_6 = 010000110$ | 14 | 13 |
| $x_7 = 010001000$ | 10 | 2 |
| $x_8 = 100000110$ | 16 | 7 |

## 5    Surface Response Estimate of Upper Bound

Experimental designs are often used to formulate a surface response equation *g(Q)* for estimating an output value. In this paper, the input parameters are the elements in the scenario Q and the output value is an estimate of *maximize $x^t Q x$*. Note that the maximum objective value is estimated, not the decision variable values. The surface response equation is a multiple linear regression that provides a very fast method of approximating the optimal value for any given input scenario when those $Q_{ij}$ values are within the bounds set by the two scenario generators. The calculated coefficients for *g*(Q) are the estimated effects of changing $Q_{ij}$ from the values defined in $Q^{upper}$ to $Q^{lower}$ and the constant term is the estimated average effect when all $Q_{ij}$ are set to their average value.

The testing from the example in section 4.1 provided a *g(Q)* with coefficients given in Table 4 with constant term 16.4. The efficacy of *g(Q)* was tested by comparing the estimate to the optimal value of 64 Q matrices whose elements are *randomly* generated between the limits set by $Q^{upper}$ and $Q^{lower}$. To calculate an upper bound with over 99% confidence, three times the standard error was added to the surface response equation estimate.

**Table 4**      Surface Response Function Coefficients

| $c_{ij}$ | $g(Q)$ coefficient | $Q^{upper}$ | $Q^{lower}$ |
|---|---|---|---|
| 0, 0 | 0.3 | -5 | 2 |
| 1, 1 | -0.1 | 2 | 2 |
| 2, 2 | -0.2 | -3 | 6 |
| 3, 3 | -0.8 | 5 | 2 |
| 4, 4 | 0.2 | -1 | 1 |
| 5, 5 | -0.5 | 2 | 0 |
| 6, 6 | -0.5 | 2 | -2 |
| 7, 7 | -0.8 | 2 | -2 |
| 8, 8 | -2.3 | 5 | -4 |
| 0, 4 | 0.3 | 1 | 3 |
| 0, 6 | -0.3 | 5 | -3 |
| 0, 7 | -0.2 | 3 | -3 |
| 0, 8 | 0.1 | 0 | -5 |
| 1, 5 | 0.0 | 6 | -1 |
| 1, 6 | -0.2 | 3 | 0 |
| 1, 7 | -0.2 | 3 | 0 |
| 2, 5 | 0.2 | 1 | 3 |
| 3, 5 | -1.5 | 7 | 1 |
| 3, 6 | 0.1 | 0 | -3 |
| 3, 7 | 0.0 | 1 | -1 |
| 3, 8 | -2.8 | 6 | -7 |
| 5, 8 | -3.0 | 9 | -5 |
| 6, 7 | -0.6 | 7 | 3 |

The $g(Q)$ estimates were then compared with those of the maximum upper bound as described in section 2. Figure 4 shows the percent each bound was above the proven optimal value and shows that $g(Q)$ consistently provided better estimates than the max upper bound. The horizontal axis is sorted from highest scenario optimality estimate gap to lowest. The bounds are determined prior to solving the problem and are not updated during the solution process. In our experience the initial relaxation of quadratic UBQP generally does not provide a good upper bound and it is not uncommon to have starting gaps of 300% or more between the best integer solution and the best initial bound. In contrast to this, the bounds generated by $g(Q)$, via the experimental design method employed here, are seen to be very respectable. As a practical matter, if the estimated upper bound on the objective value of a given scenario is below a decision maker's threshold then that scenario should be avoided.

**Figure 4**     Upper Bound Estimates Percentage of Optimal Objective Value

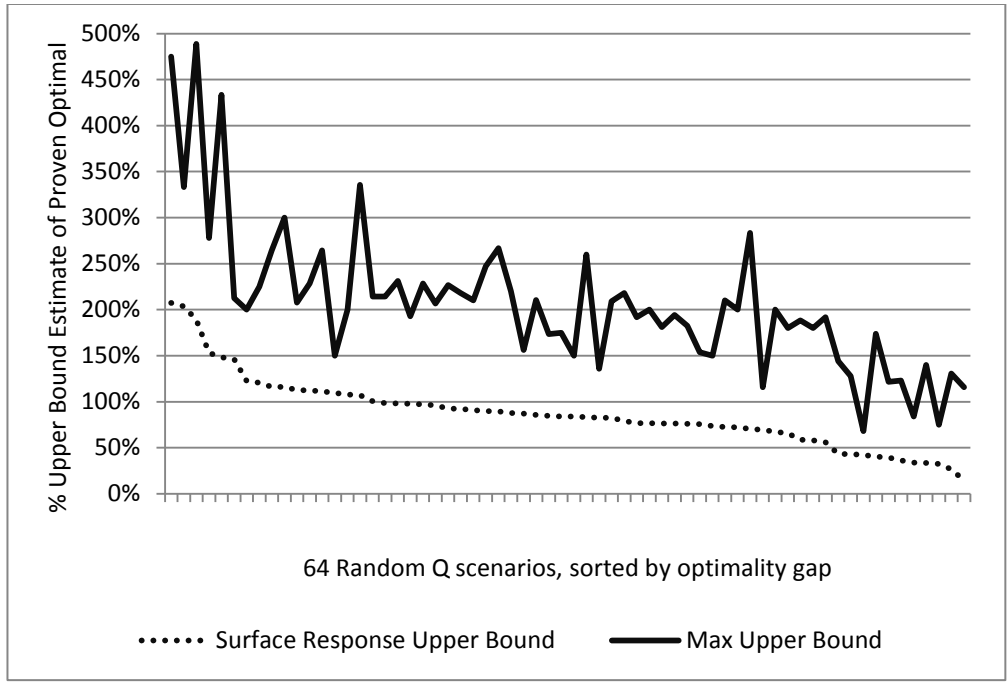

## 6     Conclusions

The unconstrained binary quadratic problem provides a framework upon which many different types of problems can be modeled.  As the UBQP framework expands into areas such as quantum annealing computers, the importance of the sensitivity, or robustness, of solutions to perturbations in the elements of Q increases.  Unique contributions to the research of robustness are an experimentally designed scenario generation and analysis approach to robust optimization yielding a new upper bound estimation based on the surface response equation.  In addition a theoretical presentation of a preprocessing approach for fixing some variables allows quantifying the robustness range over which they can be fixed, as well as the boundaries at which they become indeterminate.   The implication of fixing variables is that models involving much larger Q matrices can be reduced in size for faster scenario generation and robustness analysis.  Managerially this means that larger models based on an organisation's big data can be created and robustly analyzed.  Future work, extending what we report here, includes data mining of historical data to generate the upper and lower scenario generators and expanding the set of preprocessing rules.

# References

Alidaee, B., Kochenberger, G., & Ahmadian, A. (1994). 0-1 Quadratic programming approach for optimum solutions of two scheduling problems. *International Journal of Systems Science, 25*(2), 401-408.

Beasley, J. E. (1990). OR-Library: distributing test problems via electronic mail. *Journal of the Operational Research Society, 41*(11), 1069-1072.

Benoist, T., Estellon, B., Gardi, F., Megel, R., & Nouioua, K. (2011). LocalSolver 1.x: a black-box local-search solver for 0-1 programming. . *4OR, 9*(3), 299-316.

Bertsimas, D., Nohadani, O., & Teo, K. M. (2010). Robust Optimization for Unconstrained Simulation-Based Problems. *Operations Research, 58*(1), 161-178.

Beyer, H.-G., & Sendhoff, B. (2007). Robust optimization – A comprehensive survey. *Computer Methods in Applied Mechanics and Engineering, 196*(33-34), 3190-3218.

Billionet, A., & Elloumi, S. (2007). Using a mixed integer quadratic programming solver for the unconstrained quadratic 0-1 problem. *Math Program, 109*(1), 55-68.

Bomze, I., Budinich, M., Pardalos, P. M., & Pelillo, M. (1999). The maximum clique problem. . In *Handbook of combinatorial optimization* (pp. 1-74). Berlin: Springer.

Boros, E., & Hammer, P. (1991). The max-cut problem and quadratic 0-1 optimization polyhedral aspects, relaxations and bounds. *Annals of Operations Research, 33*(4127), 151-180.

Boros, E., Hammer, P., & Tavares, G. (2006). Preprocessing of Unconstrained Quadratic Binary Optimization. *Rutcor Research Report RRR 10-2006*, 1-54.

Choi, V. (2008). Minor-Embedding in Adiabatic Quantum Computation: I. The Parameter Setting Problem. *Quantum Information Processing,7*, 193-201.

Cohen, D. M., Dalal, S. R., Fredman, M. L., & Patton, G. C. (1997). The AETG system: an approach to testing based on combinatorial design. *IEEE Transactions on Software Engineering, 23*(7), 437-444.

Dunietz, I. S., Ehrlich, W. K., Szablak, B. D., Mallows, C. L., & Iannino, A. (1997). Applying design of experiments to software testing: experience report. *ICSE Proceedings of the 19th International Conference on Software Engineering.* Boston, MA.

Gunasekaran, A., Ho, S. C., Kwan, C.-L., & Ng, D. (in press). Hybrid Tabu Search for Effective Airport Gate Management. *International Journal of Operational Research*.

*International Collegiate Business Strategy Competition*. (2014). Retrieved 2014, from http://icbsc.org/

Joshi, K., Jain, K., & Bilolikar, V. (2016). A VNS-GA based hybrid metaheuristic for the resource constrained project scheduling problem. *International Journal of Operational Research, 27*(3), 437-449.

Kochenberger, G., Hao, J., Lu, Z., Wang, H., & Glover, F. (2013). Solving large scale max cut problems via tabu search. *Journal of Heuristics, 19*(4), 565-571.

Kochenberger, G., Hao, J.-K., Glover, F., Lewis, M., Lu, Z., Wang, H., et al. (2014). The unconstrained binary quadratic: a survey. *Journal of Combinatorial Optimization, 28*(1), 55-81.

Leger, P. M. (2006). Using a Simulation Game Approach to Teach Enterprise Resource Planning Concepts. *Journal of Information Systems Education, 17*, 441-448.

Lewis, M. (2008). On the use of Guided Design Search for Discovering Significant Decisions Variables in the Fixed charge Capacitated Multicommodity Network Design Problem. *Networks, 53*(1), 6-18.

Lewis, M., & Kochenberger, G. (2013). Graph Bisection Modeled as Cardinality Constrained Binary Task Allocation. *International Journal of Information Technology & Decision Making, 12*(2), 261-276.

Lewis, M., & Kochenberger, G. (2016). Probabilistic Multistart with Path Relinking for Solving the Unconstrained Binary Quadratic Problem. *Int Journal of Operational Research, 26*(1), 13-33.

Lewis, M., Kochenberger, G., & Alidaee, B. (2008). A new modeling and solution approach for the set-partitioning problem. *Computers & Operations Research, 35*(3), 807-813.

Libura, M. (2010). A note on robustness tolerances for combinatorial optimization problems. *Information Processing Letters , 110*(16), 725-729.

Lucas, A. (2014). Ising Formulations of Many NP Problems. *Frontiers in Physics, 5*(arXiv:1302.5843), 2.

Mulvey, J. M., Vanderbei, R., & Zenios, S. (1995). Robust Optimization of Large-Scale Systems. *Operations Research, 43*(2), 264-281.

Pan, S., Tan, T., & Jiang, Y. (2008). A global continuation algorithm for solving binary quadratic programming problems. *Computational Optimization & Applications, 41*(3), 349-362.

Samadi-Dana, S., Paydar, M., Paydar, M., & Jouzdani, J. (in press). A simulated annealing solution method for robust school bus routing. *International Journal of Operational Research*.

Wang, Y., Lu, Z., Glover, F., & Hao, J. (2012). Path relinking for unconstrained binary quadratic programming. *European Journal of Operational Research, 223*(3), 595-604.

Yilmaz, C., Dumlu, E., Cohen, M., & Porter, A. (2014). Reducing Masking Effects in Combinatorial Interaction Testing: A Feedback Driven Adaptive Approach. *IEEE Transactions on Software Engineering, 40*(1), 43-66.